\documentclass[twoside]{article}
\usepackage{amssymb}
\usepackage{amsmath}
\usepackage{algorithm}
\usepackage[noend]{algpseudocode}
\usepackage{pgfplots}
\usepackage{multicol}
\usepackage[hmarginratio=1:1,top=32mm,columnsep=20pt]{geometry}
\usepackage{fullpage}
\usepackage{pdflscape}
\usepackage[hidelinks]{hyperref}

\begin{document}
\parindent=0in
\parskip=12pt

%
%
%
%

\title{Fast MLE Computation for the Dirichlet Multinomial}

\author{Max Sklar\\
Foursquare Labs, Local Maximum Labs\\
May 2, 2014 \small{\textit{(Revision: May 26, 2023)}}
}

\date{} 

\maketitle
\thispagestyle{empty}

\begin{abstract}
Given a collection of categorical data, we want to find the parameters of a Dirichlet distribution which maximizes the likelihood of that data.  Newton's method is typically used for this purpose but current implementations require reading through the entire dataset on each iteration.  In this paper, we propose a modification which requires only a single pass through the dataset and substantially decreases running time.  Furthermore we analyze both theoretically and empirically the performance of the proposed algorithm, and provide an open source implementation.
\end{abstract}

\section{Introduction}

The Dirichlet distribution has long been used as a conjugate prior for the categorical or multinomial distribution, giving statisticians access to analytically tractable priors. Since the Dirichlet is so central to Bayesian inference, it is important to have an efficient algorithm to estimate its parameters. Here we consider the maximum likelihood estimate.

\subsection{Motivation}

One use for the Dirichlet prior is to avoid overfitting the estimate of categorical parameters from a small amount of data.  For example, consider a recommendation engine that aggregates user-generated ratings. With only a few users weighing in on a given item, the maximum likelihood estimate of the rating distribution can be very inaccurate. The Dirichlet prior prevents the estimate from being overfit when only one or two users have left an opinion.

In an example of higher dimension, the relative popularity of a venue on Foursquare at a given time of week is estimated using a Dirichlet prior\cite{sklar}.  The week is divided in to $168$ buckets, each representing one hour in the week.  The timeliness distribution of a venue starts out as a Dirichlet prior, and is updated whenever new visit data is received. The Dirichlet prior is originally derived from the visit data of many similar Foursquare venues using the algorithm described in this paper.

Of course, not all applications of the Dirichlet prior are new.  In the 1960s, James Mosimann found applications in analyzing biological data\cite{mosimann}.  In one example, there were 4 types of pollen found in pollen cores at different levels.  For each core, only $100$ grains of pollen were counted, giving relatively sparse data for any given core.  However, even with only 73 cores analyzed, a lot more could be said about the data after modeling it as a Dirichlet-multinomial\cite[pg 8]{ng}\cite{mosimann}.

A Dirichlet distribution is much simpler to estimate if the data comes in the form of categorical distribution parameters drawn from that Dirichlet. But what if we only have access to a few samples from each categorical distribution?  In fact, if we already had the categorical parameters, or had sampled it so heavily that the maximum likelihood estimate is accurate, there would be no need for a Dirichlet prior in the first place.

Perhaps the largest motivation for this effort is that the Dirichlet and Dirichlet-multinomial distributions can be used as part of a larger system. Consider that the K-means algorithm cannot perform well without an efficient calculation for the mean of a set of points. Likewise, a mixture of Gaussians cannot be computed without first computing one Gaussian.  If one intends to build a mixture model of Dirichlet distributions by using the Expectation-Maximization algorithm, having an efficient way to estimate Dirichlet parameters is valuable.

Dirichlet distributions are also widely used in topic models for Natural Language Processing.  For example, the latent Dirichlet allocation requires a computation of the MLE for a Dirichlet\cite{blei}\cite{heinrich}\cite{wallach}.  As the datasets grow larger and the models more complex, the demand for computational efficiency increases.  

\subsection{Previous Work}

In 1989, Gerd Ronning published a paper on estimating a Dirichlet from a set of multinomial samples\cite{ronning}. Ronning provides a proof that the objective function is globally concave and therefore the Newton-Raphson method will converge to a solution \cite[pg 73]{ng}.  In 1990, Narayanan published a paper on the MLE computation of the Dirichlet including an implementation in Fortran\cite{narayanan}. He cites various examples of Dirichlet distributions being used for statistical models going back to the 1960s and 1970s, including Mosiman's work on serum proteins in white Pekan ducklings\cite[pg 8]{ng}.

In 2000, Thomas Minka\cite{minka} wrote about the MLE for Dirichlet distributions, and later wrote an implementation in Matlab.

Much work has also been done in finding an initialization for the Newton-Raphson method \cite[pg 74-54]{ng}.  In 1980, Dishon and Weiss\cite{dishon} proposed using the simple method of moments for the Beta distribution \footnote{the 2-dimensional case of the dirichlet distribution}.  In 1989, Ronning\cite{ronning} produced a new initialization and proved that it was guaranteed to stay within certain bounds for the first Newton-Raphson step.  In 2003, Hariharan and Vela\cite{hariharan} proposed a new, simpler, method for initialization.  And finally in 2008, Wicker et al\cite{wicker}, proposed a maximum likelihood approximation method which was shown to work more efficiently with some standard data sets.

All of that work assumes direct access to the multinomial parameters, and not from just count data.  Minka\cite{minka} tackled this problem in 2000 and describes the computations for finding the MLE of a Dirichlet multinomial.   His algorithms include the Newton-Raphson method and a fixed-point iteration.  He also includes a Matlab implementation.  In 2008, Hanna Wallach reviewed Minka's algorithms, and developed a new fixed-point iteration based on a mathematical transformation of the digamma function leading to a precomputation.  Ng et al\cite{ng} in 2011 also describe the mathematics, and say that ``in general, Newton's method provides the fastest means for finding the MLE''.

This paper builds on the work of Minka and Wallach to compute the MLE of a Dirichlet-multinomial distribution.  The proposed algorithm uses the Newton-Raphson method, and arrives at the same value as other MLE methods\footnote{In fact, given the same initial conditions, every iteration of Newton-Raphson is untouched.}.  It takes advantage of a precomputation that can be made using the properties of the digamma and trigamma function, not unlike Wallach's approach in the fixed-point case.  The proposed algorithm is faster in a wide range of circumstances, particularly in the case where count data for each multinomial instance is scarce.

\section{Definitions and Conventions}

The \textit{gamma function} is a positive-valued function whose domain is the positive real numbers.\footnote{It is typically generalized to complex numbers but only positive values are considered in this paper.}

\[\Gamma(x)=\int_0^\infty t^{x+1}e^{-x}dx\]

The gamma function of a positive integer n is equivalent to n minus 1 factorial:

\[\Gamma(n)=(n-1)!\]

We define the dual-input gamma function of a positive number and a natural number \(n\) as the ratio of the gamma function of \(a + n\) and \(a\). This is also called a \textit{rising power}\cite{concrete} \(a^{\bar{n}}\).

\[\Gamma(a,n)=\frac{\Gamma(a+n)}{\Gamma(a)}=\prod_{i=0}^{n-1}(a+i)=a^{\bar{n}}\]

In order to simplify the expressions in this paper, we also introduce the dual-input log-gamma function as equivalent to the log of \(\Gamma(a, n)\)

\[L\Gamma(a,n)=\sum_{i=0}^{n-1}\ln(a+i)\]

A \textit{\(K\)-dimensional categorical distribution} or \textit{general finite distribution}\cite{ng} is a probability distribution over K mutually exclusive outcomes.  A categorical distribution is represented by a \(K\)-dimensional vector $\mathbf{p}$, where all the values sum to 1.

\[\sum_{k=0}^{K-1}p_k=1\]

Each sample from a categorical distribution is a member of \(\{0, 1, 2, ..., K-1\}\). When this distribution is sampled \(n\) times, the results can be summarized as a K-dimensional count vector \(\mathbf{c}\), where \(c_k\) is the number of times category \(k\) appears in the data set. The counts all sum to $n$.

\[\sum_{k=1}^{K-1}c_k=n\]

The probability distribution over all possible count vectors is called a \textit{multinomial distribution}, or more precisely in this case, a \textit{K-dimensional multinomial distribution with n trials}. Given a categorical distribution and \(n\), the resulting multinomial distribution can be computed precisely using - appropriately - \textit{multinomial coefficients}.

A \textit{\(K\)-dimensional Dirichlet distribution} is a family of probability distributions over the space of K-dimensional categorical distributions. It is parameterized by a K-dimensional vector $\alpha$, and in terms of $\mathbf{p}$ it is proportional to $\prod_{k=0}^{K-1}{p_k}^{\alpha_k-1}$. With a normalization term, the distribution function for the Dirichlet is

\begin{equation}
\label{eq:dirich_distr}
\mathcal{D}(\mathbf{p},\mathbf{\alpha})=\frac{\prod_{k=0}^{K-1}\Gamma(\alpha_i)}{\Gamma\left(\sum_{k=0}^{K-1}\alpha_k\right)}\prod_{k=0}^{K-1}{p_k}^{\alpha_k-1}.
\end{equation}

Suppose that a count vector of \(n\) samples is drawn from an unknown categorical distribution. If the bayesian prior for this distribution is a Dirichlet, then the posterior is also a Dirichlet. This makes the Dirichlet distribution a \textit{conjugate prior} for the categorical distribution.

More specifically, if \(\mathcal{D}(\mathbf{p},\mathbf{\alpha})\) represents our prior probability distribution over \(\mathbf{p}\) and \(\mathbf{c}\) is the count vector for the observed sample, then the posterior distribution is $P(\mathbf{p} | \mathbf{c})=\mathcal{D}(\mathbf{p},\mathbf{\alpha} + \mathbf{c})$. This result is satisfyingly elegant and makes Bayesian inference with the Dirichlet distribution and count data far simpler than one would expect.

The expected value of a Dirichlet distribution is obtained by normalizing the vector $\alpha$.

\[\mathbb{E}(\mathcal{D}(\mathbf{p},\mathbf{\alpha}))=\frac{\mathbf{\alpha}}{\sum_{i=0}^{K-1}\alpha_i}\]

An intuitive notion of the Dirichlet distribution is that each component of $\alpha$ puts some weight on each category before the data is collected. As more samples are obtained, the prior value of $\alpha$ becomes less and less relevant and the data is allowed to speak for itself.

The probability distribution over the possible configurations of a count vector given a Dirichlet distribution is the \textit{Dirichlet-multinomial distribution}.  It is derived as follows:

\begin{equation}
P(\mathbf{c}|\mathbf{\alpha})=\int_{\mathbf{p}}P(\mathbf{c}|\mathbf{p})P(\mathbf{p}|\mathbf{\alpha})=
\frac{\prod_{k=0}^{K-1}\Gamma(\alpha_k,c_k)}{\Gamma\left(\sum_{k=0}^{K-1}\alpha_k,\sum_{k=0}^{K-1}c_k\right)}
\frac{\Gamma\left(1,\sum_{k=0}^{K-1}c_k\right)}{\prod_{k=0}^{K-1}\Gamma(1,c_k)}
\label{eq:dirichlet_multinomial_formula}
\end{equation}

Finally, the \textit{maximum likelihood estimate (MLE)} for a parameterized probability distribution are those parameters which maximize the likelihood of the data.

\[MLE(\mathbf{\alpha})=\arg\max_\mathbf{\alpha}\ P(\mathbf{c}|\mathbf{\alpha})\]

\section{Overview of the Problem}

Suppose that $N$ samples are drawn from a Dirichlet distribution, and we want to find the MLE of the Dirichlet parameter $\alpha$.  There are two cases to consider.

In the Dirichlet problem, we know the exact probabilities of the categorical distribution for each sample. We illustrate this type of dataset in Table~\ref{tab:d}. Each row in the table represents a sample and each column represents a category. Each individual value in the table displays the probability of the corresponding category for the corresponding sample.

\begin{table}[h] \centering
\begin{tabular}{ l | l l l | l}
  sample & \(k=0\) & \(k=1\) & \(k=2\) & sum \\ \hline
  0 & 0.21 & 0.37 & 0.42 & 1\\
  1 & 0.56 & 0.15 & 0.29 & 1\\
  \vdots & \vdots & \vdots & \vdots & \vdots\\
  \(n-1\) & 0.14 & 0.33 & 0.53 & 1\\
\end{tabular}
\caption{Dataset in the ``Pure'' Dirichlet Problem}\label{tab:d}
\end{table}

In the \textit{Dirichlet-multinomial problem}, we cannot directly observe the probabilities associated with each category. Instead, we record a count vector by sampling each row. The data is going to look more like Table~\ref{tab:dm} where each value represents a count rather than a probability.

\begin{table}[h] \centering
\begin{tabular}{ l | l l l | l}
  sample & \(k=0\) & \(k=1\) & \(k=2\) & sum \\ \hline
  0 & 3 & 14 & 0 & 17\\
  1 & 1 & 16 & 3 & 20\\
  \vdots & \vdots & \vdots & \vdots & \vdots\\
  \(n-1\) & 6 & 3 & 10 & 19\\
\end{tabular}
\caption{Dataset in the Dirichlet-Multinomial Problem}\label{tab:dm}
\end{table}

In both cases, the Newton-Raphson method can be used to accurately compute the MLE\cite{minka}.  The pure Dirichlet case is relatively fast because a \textit{sufficient statistic} compressing all the data into a single \(K\)-dimensional vector can be precomputed.  Unfortunately, the Dirichlet-multinomial problem cannot be reduced in the same way, making it an order of magnitude slower.

The logical improvement is therefore to find an alternative pre-computation on the Dirichlet-multinomial data to speed up the algorithm.  Instead of compressing the data into a \(K\)-dimensional vector, the Dirichlet-multinomial data can in fact be compressed into a \(K\) x \(M\) matrix, and an \(M\)-dimensional vector, where \(M\) is the maximum sum of any row.

\section{The Fast Dirichlet Solution}
\label{fastdirichlet}

This is equivalent to the solutions documented by Minka\cite{minka} (2000), Blei\cite[A.4.2]{blei} (2003) and Ng\cite[pg 72-73]{ng} (2011).  The dataset for the Dirichlet problem is an \(N\) x \(K\) matrix $D$ in which each row represents a sample from a single Dirichlet.  Let $d_{n,k}$ represent the nth row and kth column of matrix $D$.  Given a Dirichlet parameter $\alpha$, we can use equation~\eqref{eq:dirich_distr} to find the probability of the dataset as
\[
P(D|\alpha)=\prod_{n=0}^{N-1}\Gamma\left(\sum_{k=0}^{K-1}\alpha_k\right)\prod_{k=0}^{K-1}\Gamma(\alpha_k)^{-1}d_{n,k}^{\alpha_k-1}.
\]
We want to maximize this probability with respect to $\alpha$.  One way to do this is to build and maximize a function $F$, which is related to the log of the probability.
\[
\ln(P(D|\alpha))=\sum_{n=0}^{N-1}\ln\Gamma\left(\sum_{k=0}^{K-1}\alpha_k\right) + \sum_{n=0}^{N-1}\sum_{k=0}^{K-1}\left[-\ln\Gamma(\alpha_k) + (\alpha_k-1)\ln\left(d_{n,k}\right)\right]
\]
We can distribute the summations and remove the constant terms.\footnote{Note that we divided the entire equation by $N$.  The number of samples does not factor in to the MLE, but it does control the variance around that value.}
\[
F(\alpha)=\ln\Gamma\left(\sum_{k=0}^{K-1}\alpha_k\right) - \sum_{k=0}^{K-1}ln\Gamma(\alpha_k) + \sum_{k=0}^{K-1}\alpha_k\left(\frac{1}{N}\sum_{n=0}^{N-1}\ln\left(d_{n,k}\right)\right)
\]

In this form, only one term involves the data. This term is called the \textit{sufficient statistic} and will be set as the vector $\mathbf{v}$. The sufficient statistic can be computed with one pass through the data.

\[v_k=\frac{1}{N}\sum_{n=0}^{N-1}\ln\left(d_{n,k}\right)\]

The distinction separating the Dirichlet and the Dirichlet-multinomial cases is that the Dirichlet case has a sufficient statistic of constant dimension.  In fact, this is true for all families of exponential distributions\cite[pg 116]{robert}.  The function to maximize now looks like this:

\[
F(\alpha)=\ln\Gamma\left(\sum_{k=0}^{K-1}\alpha_k\right)-\sum_{k=0}^{K-1}\ln\Gamma\left(\alpha_k\right)+\sum_{k=0}^{K-1}\alpha_kv_k
\]

In order to use the Newton-Raphson method, we need to compute the gradient vector $\mathbf{g}$ and the Hessian matrix $H$.  The formula for a single step is as follows:

\[\alpha_{new}=\alpha_{old}-H^{-1}\mathbf{g}\]

The values for $H$ and $\mathbf{g}$ also require only the vector $\mathbf{v}$, and not the full dataset.

We will use the following derivatives:
\[
\begin{array}{cc} \Psi(x) = \frac{d}{dx} \ln\Gamma(x) & \Psi'(x) = \frac{d}{dx} \Psi(x)
\end{array}
\]
These functions, along with gamma are available in most numeric libraries.  The Python code linked to this paper uses the Numpy library.  In the past, Narayanan\cite{narayanan} and Minka\cite{minka} relied on Fortran and Matlab implementations respectively.

The gradient is given as follows:
\[
g_k=\Psi\left(\sum_{k=0}^{K-1}\alpha_k\right)-\Psi(\alpha_k)+v_k
\]

The Hessian matrix consists of a constant matrix, plus a diagonal.  Let $\mathbb{I}_{i,j}$ be the indicator function that is equal to $1$ if $i = j$ and $0$ otherwise.

\[
h_{i,j}=\Psi'\left(\sum_{k=0}^{K-1}\alpha_k\right)-\Psi'(\alpha_i)\mathbb{I}_{i,j}
\]

There is a formula for efficiently inverting this type of matrix, and it is a special case of the matrix inversion lemma\cite{woodbury}.  The technique is also used by Minka\cite{minka} (2000), Blei\cite[A.2]{blei}(2003), and Ng\cite[214]{ng} for Dirichlet-based machine learning. Let the vector representing the diagonal be $\mathbf{d}$ where $\mathbf{d}_k = -\Psi'(\alpha_k)$, and the scalar constant in the constant matrix be $c = \Psi'\left(\sum_{k=0}^{K-1}\alpha_k\right)$.  The formula computing the inverse Hessian is

\begin{equation}
H=\operatorname{diag}(\mathbf{d}) + \mathbf{1}\mathbf{1}^Tc
\qquad \qquad
H^{-1}=\operatorname{diag}(\mathbf{d})^{-1}-\frac{\operatorname{diag}(\mathbf{d})^{-1}\mathbf{1}\mathbf{1}^T\operatorname{diag}(\mathbf{d})^{-1}}{c^{-1}+\sum_{k=0}^{K-1}d_k^{-1}}.
\label{eq:invert_hessian}
\end{equation}

The process for turning this matrix inversion into an algorithm has been documented in \textit{Algorithms for Multivariate Newton-Raphson for Optimization}\cite{addendum} that serves as an addendum to this paper. The result is the following methods that can be broadly implemented in a programming language.

\begin{algorithm}
  \caption{Algorithm for one Newton Step when the Hessian Matrix is a Constant plus a Diagonal}
  \begin{multicols}{2}
  \begin{algorithmic}
  \Function{Step}{$\mathbf{g} \in \mathbb{R}^{K},\mathbf{d} \in \mathbb{R}^{K},c \in \mathbb{R}$}
  \State $S \leftarrow 0$
  \For  {$k=0$ to $K-1$}
    \State $S \leftarrow S + \mathbf{g}[k] / \mathbf{d}[k]$
  \EndFor
  \State $Z \leftarrow 1/c$
  \For  {$k=0$ to $K-1$}
    \State $Z \leftarrow Z + 1 / \mathbf{d}[k]$
  \EndFor
  \State $\mathbf{\delta} \in \mathbb{R}^{K}$
  \For  {$k=0$ to $K-1$}
    \State $\mathbf{\delta}[k] \gets (S/Z - \mathbf{g}[k]) / \mathbf{d}[k]$
  \EndFor
  \State \Return $\mathbf{\delta}$
  \EndFunction
  \end{algorithmic}
  \columnbreak
  \begin{algorithmic}
  \Function{StepInLogSpace}{$\alpha \in \mathbb{R}^{K},\mathbf{g},\mathbf{d},c$}
  \State $x \in \mathbb{R}^{K}$
  \For  {$k=0$ to $K-1$}
    \State $x[k] \gets \mathbf{g}[k]+\mathbf{\alpha}[k] \cdot \mathbf{d}[k]$
  \EndFor
  \State $Z \gets 1/c$
  \For  {$k=0$ to $K-1$}  
    \State $Z \gets Z + \mathbf{\alpha}[k] / \mathbf{x}[k]$
  \EndFor
  \State $S \gets 0$
  \For  {$k=0$ to $K-1$}  
    \State $S \gets S + \mathbf{\alpha}[k] \cdot \mathbf{g}[k]/\mathbf{x}[k]$
  \EndFor
  \State $\mathbf{\delta} \in \mathbb{R}^{K}$
  \For  {$k=0$ to $K-1$}  \State $\mathbf{\delta}[k] \gets (S / Z - \mathbf{g}[k]) / \mathbf{x}[k]$ \EndFor
  \State \Return $\delta$
  \EndFunction
  \end{algorithmic}
  \end{multicols}
  \label{alg:step}
\end{algorithm}

Algorithm~\ref{alg:step} computes one Newton step given the gradient $\mathbf{g}$, the Hessian diagonal vector $\mathbf{d}$, and the Hessian constant $c$. It is derived from equation~\eqref{eq:invert_hessian}. The alternative \textbf{StepInLogSpace} does the same computation on the vector \([ln(\alpha_k)]\) to guarantee that all \(\alpha_k\) values remain positive. \textbf{StepInLogSpace} can be used at any time, but particularly when the original \textbf{Step} function takes any \(\alpha_k\) into negative territory.

\section{The Dirichlet-Multinomial Solution}

The technique for the fast Dirichlet-multinomial solution is similar. The data is still an $N$ x $K$ matrix, except now each entry is a count, representing the number of times that category has been seen while sampling that row. Typically some rows will have more samples than others.  If a row has all zeros, no data was gathered from this sample, and it will not affect the estimate of the Dirichlet.

We can use the Dirichlet-multinomial formula in equation~\eqref{eq:dirichlet_multinomial_formula} to find the probability of seeing the data matrix $D$ given Dirichlet distribution \(\alpha\).
\[
P(D|\alpha)= \prod_{n=0}^{N-1}\prod_{k=0}^{K-1}\Gamma\left(\alpha_k,d_{n,k}\right)\cdot\Gamma\left(\sum_{k=0}^{K-1}\alpha_k,\sum_{k=0}^{K-1}d_{n,k}\right)^{-1}\cdot\Gamma\left(1,\sum_{k=0}^{K-1}d_{n,k}\right)\cdot\prod_{k=0}^{K-1}\Gamma\left(1,d_{n,k}\right)^{-1}
\]

We now apply the logarithm to simplify the process of maximizing the likelihood.
\[
\ln(P(D|\alpha))= \sum_{n=0}^{N-1}\left(\sum_{k=0}^{K-1}L\Gamma(\alpha_k,d_{n,k}) -L\Gamma\left(\sum_{k=0}^{K-1}\alpha_k,\sum_{k=0}^{K-1}d_{n,k}\right)+ L\Gamma\left(1,\sum_{k=0}^{K-1}d_{n,k}\right)-\sum_{k=0}^{K-1}L\Gamma(1,d_{n,k})\right)
\]
The last two terms are constant with respect to $\alpha$ and can be omitted in the maximization function.
\begin{equation} \label{eq:original_dmf}
F(\alpha)=\sum_{n=0}^{N-1}\sum_{k=0}^{K-1}L\Gamma(\alpha_k,d_{n,k})-
\sum_{n=0}^{N-1}L\Gamma\left(\sum_{k=0}^{K-1}\alpha_k,\sum_{k=0}^{K-1}d_{n,k}\right)
\end{equation}
The computation of $F$ requires a full read through the data matrix for every value of \(\alpha\). Because the probability distribution is not in exponential form, there cannot be a sufficient statistic of constant dimension\cite[pg 116]{robert}. Fortunately, a sufficient statistic of variable dimension does exist. We start to find this by expanding the formula for \(L\Gamma\) and rearranging the terms.
\begin{equation} \label{eq:fulldmf}
F(\alpha)= \sum_{n=0}^{N-1}\sum_{k=0}^{K-1}\sum_{i=0}^{d_{n,k}-1}\ln\left(\alpha_k+i\right)-
\sum_{n=0}^{N-1}\sum_{i=0}^{\Sigma_kd_{n,k}-1}\ln\left(\sum_{k=0}^{K-1}\alpha_k+i\right)
\end{equation}
Observe that all of the terms in equation~\eqref{eq:fulldmf} have a similar form.  In the first summation block, they each involve the log of an alpha plus a count.  In the second block, it’s the sum of the alphas plus a count. Since some of these terms may appear more than once, so it’s natural to gather them.  Let $M$ be the highest possible value of the index $i$, which will be equal to the highest row total.
\[M = \max_n\Sigma_kd_{n,k}\]
We simplify the formula for $F$ by constructing an $M$-dimensional vector $\mathbf{v}$, and a $K$ x $M$ dimensional matrix, $U$.  Let $u_{k,m}$ index the matrix $U$, and $v_m$ index the vector $\mathbf{v}$.
\begin{equation} \label{eq:finaldmf}
F(\alpha)=\sum_{k=0}^{K-1}\sum_{m=0}^{M-1}u_{k,m}\ln(\alpha_k+m)-
\sum_{m=0}^{M-1}v_m\ln(\Sigma_k\alpha_k+m)
\end{equation}

Intuitively, the values $u_{k,m}$ represent the number of rows in the data where the count in column $k$ exceeds $m$.  The value $v_m$ represents the number of rows in the data where the total sample in the row exceeds $m$.  
\[
u_{k,m}=\sum_{n=0}^{N-1}\mathbb{I}_{d_{n,k} > m} \qquad v_m=\sum_{n=0}^{N-1}\mathbb{I}_{\sum_{k=0}^{K-1}d_{n,k} > m}
\]
\begin{algorithm}
  \caption{Precomputation: Computing $U$ and $\mathbf{v}$ from the dataset}
  \begin{algorithmic}
  \Function{PrecomputeUV}{$D \in \mathbb{R}^{N \times K}$, $M \in \mathbb{N}$}
    \State $u \gets \text{zeros}(K, M)$
    \State $v \gets \text{zeros}(M)$
    \For{$n \gets 0$ \textbf{to} $N - 1$}
      \State $m \leftarrow 0$
      \For {$k=0$ to $K-1$}
        \For {$i=0$ to $D[n][k]$}
          \State $u[k][i] \leftarrow u[k][i] + 1$
          \State $v[m] \leftarrow v[m] + 1$        
          \State $m \leftarrow m + 1$
        \EndFor
      \EndFor
    \EndFor
    \State \Return $u, v$
  \EndFunction
  \end{algorithmic}
  \label{alg:uv}
\end{algorithm}

Algorithm~\ref{alg:uv} performs the key pre-computation of $U$ and $\mathbf{v}$. To obtain the gradient of F, we can now use the formula
\[
g_k= \sum_{m=0}^{M-1}u_{k,m}(\alpha_k+m)^{-1}-
\sum_{m=0}^{M-1}v_m(\Sigma_k\alpha_k+m)^{-1}.
\]
Fortunately, the Hessian matrix below in equation~\eqref{eq:hessian_dm} is still a constant plus a diagonal.  Algorithm~\ref{alg:step} can be applied to the results to perform a Newton-Raphson step.
\begin{equation} \label{eq:hessian_dm}
h_{i,j}=\sum_{m=0}^{M-1}v_m(\Sigma_k\alpha_k+m)^{-2}-
\sum_{m=0}^{M-1}u_{i,m}(\alpha_i+m)^{-2}\mathbb{I}_{i,j}
\end{equation}
\section{Theoretical Runtime Analysis}
In this section, we analyze the running time for the proposed algorithm, and compare it to Minka's version.
\subsection{Minka's Algorithm}
If the function $F$ were calculated from the summations in equation~\eqref{eq:fulldmf}, then the running time would be $O(MNK)$.  However, algorithms exist for the log-gamma, digamma, and trigamma functions which are more efficient than simply adding up the logarithmic terms\cite{wallach}. We assume that these can be run in constant time.  From the double summation in equation~\eqref{eq:original_dmf}, the running time becomes $O(NK)$, or the size of the data set.

If the $F$ can be computed in $O(NK)$, so can the gradient and the Hessian components needed to run algorithm~\ref{alg:uv}. If the total number of Newton steps needed to find the MLE within a given precision is $s$, the final running time is $O(sNK)$.
\subsection{Proposed Algorithm}

Algorithm~\ref{alg:uv} describes the initial sweep through the dataset to calculate $U$ and $\mathbf{v}$.  This will have a running time of $O(MNK)$.
For the maximization function, we are now looking at equation~\eqref{eq:finaldmf}.  The calculation of this function is $O(MK)$ and therefore so is the calculation of the gradient, the Hessian, and the entire Newton-Raphson step from algorithm~\ref{alg:step}.

Again assuming that we require $s$ steps in Newton's Method, the running time for that phase of the calculation will be of $O(sMK)$.
Putting it together, the calculation of Newton's method plus the precomputation of $U$ and $\mathbf{v}$ is $O(NMK + sMK)$ or $O((N+s)MK)$.

If $s$ is an order of magnitude lower than $N$, then the bottleneck is reading in the data, which makes the running time effectively $O(NMK)$.\footnote{Fortunately, $U$ and $\mathbf{v}$ are additive, so this processes can be parallelized across multiple processors in map-reduce job. More about this in the future work section.} When the Newton updates are the bottleneck, the algorithm will have a running time of $O(sMK)$.  In that case, the proposed algorithm does not have a running-time benefit over Minka's version.

\section{Running Time Experiments} \label{section:experiments}
The evaluation of the proposed algorithm uses the Fastfit library\footnote{http://research.microsoft.com/en-us/um/people/minka/software/fastfit/}, written by Thomas Minka at Microsoft Research\cite{minka}.  Fastfit is a Matlab toolbox and contains two functions for evaluating a Dirichlet-multinomial. A third function based on the proposed algorithm was written in order to compare running times on the same platform (See section \ref{code}).  The initialization and stopping criteria were left untouched in order to make a fair comparison\footnote{This implementation stops when the absolute value of the gradient is less than $10^{-16}$}.

According to the runtime analysis, the proposed algorithm will perform very well when M is small and N grows large.  Samples were generated with a 3-dimensional Dirichlet parameter $[3,1,2]$, and a constant $M = 10$.  The value for $N$ starts at $100$ and doubles on each round.

This analysis also includes Minka's fixed-point (FP) solution and a new Matlab version using $U$ and $\mathbf{v}$.  The fixed-point algorithm in Fastfit is
\[
\alpha_k^{*}= \alpha_k\frac{\sum_{n=0}^{N-1}\Psi(d_{n,k}+\alpha_k)-\Psi(\alpha_k)}{\sum_{n=0}^{N-1}\Psi(\sum_{k=0}^{K-1}d_{n,k}+\alpha_k)-\Psi(\sum_{k=0}^{K-1}\alpha_k)}.
\]

The proposed version makes use of the parameters $U$ and $\mathbf{v}$ and has the advantage of not having to run through the dataset multiple times.
\[
\alpha_k^{*}= \alpha_k\frac{\sum_{m=0}^{M-1}u_{k,m}(\alpha_k+m)^{-1}}{\sum_{m=0}^{M-1}v_m\left(\sum_{k=0}^{K-1}\alpha_k+m\right)^{-1}}
\]

This is equivalent to Wallach's algorithm using the digamma recurrence relation\cite{wallach}. The Fastfit fixed-point algorithm has an advantage over the Fastfit Newton algorithm because the bottleneck is reading the data on each iteration, and Newton's method requires more computations per data point.  While Newton's method requires computation of both the gradient and the Hessian, the fixed point algorithm needs only the terms from the gradient.  With the precalculation of $U$ and $\mathbf{v}$, that advantage for fixed-point iteration becomes less important.

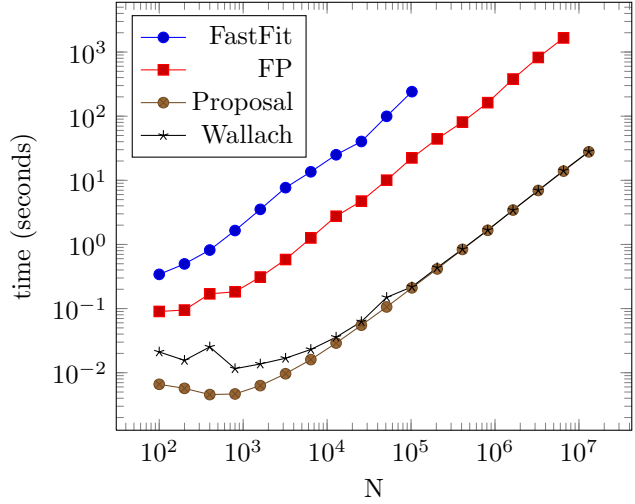
\begin{center}
\begin{tikzpicture}
\begin{loglogaxis}[
xlabel=N,
ylabel=time (seconds),
legend style={
cells={anchor=east},
legend pos= north west,
}]

\addplot coordinates {
(100,0.3414)
(200,0.49696)
(400,0.81785)
(800,1.6474)
(1600,3.5246)
(3200,7.6819)
(6400,13.5708)
(12800,25.0067)
(25600,40.2966)
(51200,99.487)
(102400,240.9979)
};
\addlegendentry{FastFit}

\addplot coordinates {
(100,0.090174)
(200,0.094939)
(400,0.16981)
(800,0.18266)
(1600,0.31075)
(3200,0.58193)
(6400,1.2623)
(12800,2.7547)
(25600,4.7415)
(51200,10.047)
(102400,22.3356)
(204800,44.2398)
(409600,81.0067)
(819200,162.5518)
(1638400,379.775)
(3276800,822.923)
(6553600,1658.4084)
};
\addlegendentry{FP}

\addplot coordinates {
(100,0.006598)
(200,0.005703)
(400,0.004562)
(800,0.004662)
(1600,0.006295)
(3200,0.009634)
(6400,0.016019)
(12800,0.028998)
(25600,0.055058)
(51200,0.10605)
(102400,0.21035)
(204800,0.41523)
(409600,0.83691)
(819200,1.6634)
(1638400,3.4372)
(3276800,6.9482)
(6553600,13.9044)
(13107200,27.7929)
};
\addlegendentry{Proposal}

\addplot coordinates {
(100,0.021078)
(200,0.015583)
(400,0.025232)
(800,0.011562)
(1600,0.013674)
(3200,0.016818)
(6400,0.023057)
(12800,0.035835)
(25600,0.06255)
(51200,0.14943)
(102400,0.21851)
(204800,0.43566)
(409600,0.84666)
(819200,1.6761)
(1638400,3.4277)
(3276800,6.9522)
(6553600,14.1749)
(13107200,28.0573)
};
\addlegendentry{Wallach}

\end{loglogaxis}
\end{tikzpicture}
\end{center}

The running time for increasing $N$ on the proposed algorithm is hinge-shaped; there is a section where it is constant, and a section where it is increasing linearly.  This is in agreement with the running time $O(NK+ sMK)$.  The constant section is dominated by the term $O(sMK)$, and the linear section by the term $O(NK)$.  While any MLE algorithm which reads all of the data has to be at least linear time for increasing $N$, the proposed algorithm reduces running time by a constant factor of over $1000$ in this case.  For high $N$, Algorithm~\ref{alg:uv} dominates the running time for both the proposed algorithm and Wallach's algorithm.

Since the proposed algorithm divides into two parts, the next graph compares the time for precomputation with the time for executing Newton-Raphson.

\begin{center}
\begin{tikzpicture}
\begin{loglogaxis}[
xlabel=N,
ylabel=time (seconds),
legend style={
cells={anchor=east},
legend pos= north west,
}]

\addplot coordinates {
(200,0.000449)
(400,0.000846)
(800,0.001632)
(1600,0.003222)
(3200,0.006382)
(6400,0.012776)
(12800,0.025522)
(25600,0.051014)
(51200,0.10199)
(102400,0.21952)
(204800,0.40879)
(409600,0.81845)
(819200,1.6373)
(1638400,3.2788)
(3276800,6.5526)
(6553600,13.0938)
(13107200,26.2593)
(26214400,52.569)
};
\addlegendentry{Precomputation}

\addplot coordinates {
(200,0.004302)
(400,0.003408)
(800,0.002818)
(1600,0.00277)
(3200,0.002773)
(6400,0.002798)
(12800,0.002803)
(25600,0.002792)
(51200,0.002824)
(102400,0.002881)
(204800,0.003033)
(409600,0.003293)
(819200,0.003069)
(1638400,0.003064)
(3276800,0.002816)
(6553600,0.003071)
(13107200,0.003051)
(26214400,0.003048)
};
\addlegendentry{N-R Learning}

\end{loglogaxis}
\end{tikzpicture}
\end{center}

When $N$ gets large enough, the running time of the precomputation becomes much larger than the running time of the Newton-Raphson phase. To provide increasingly accurate intermediate answers as the data is processed, we periodically run Newton-Raphson before all the rows are processed without adding significant overhead. This allows for an implicit online learning mode, where the model's accuracy is progressively improved as new data is received.

\subsection{Limitations of the Proposed Algorithm}

When $M$ is large and $N$ is held constant, the proposed algorithm is not expected to run as well.  Here the same Dirichlet parameter was used to generate the dataset and $N$ was held constant at $5000$.

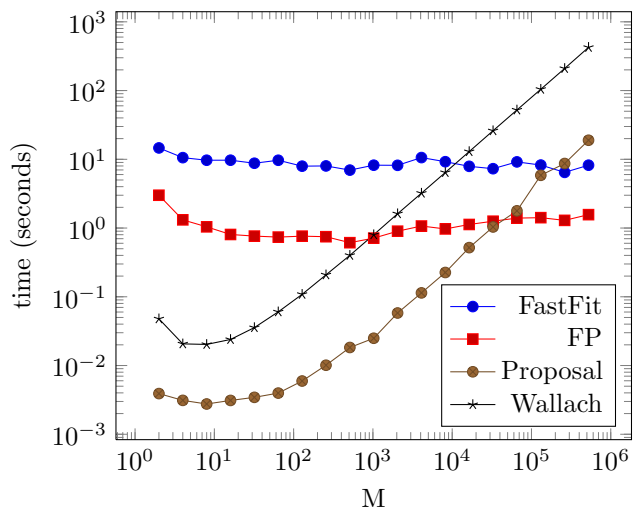
\begin{center}
\begin{tikzpicture}
\begin{loglogaxis}[
xlabel=M,
ylabel=time (seconds),
legend style={
cells={anchor=east},
legend pos= south east,
}]

\addplot coordinates {
(2,14.6338)
(4,10.5691)
(8,9.6943)
(16,9.703)
(32,8.7648)
(64,9.6872)
(128,7.9455)
(256,8.0075)
(512,6.963)
(1024,8.2081)
(2048,8.1749)
(4096,10.5937)
(8192,9.224)
(16384,7.904)
(32768,7.298)
(65536,9.1676)
(131072,8.2438)
(262144,6.4546)
(524288,8.2083)
};
\addlegendentry{FastFit}

\addplot coordinates {
(2,3.0017)
(4,1.3149)
(8,1.0411)
(16,0.80595)
(32,0.76357)
(64,0.73871)
(128,0.76202)
(256,0.74796)
(512,0.6122)
(1024,0.71341)
(2048,0.89982)
(4096,1.0698)
(8192,0.97032)
(16384,1.1267)
(32768,1.257)
(65536,1.3945)
(131072,1.4157)
(262144,1.2933)
(524288,1.566)
};
\addlegendentry{FP}

\addplot coordinates {
(2,0.003906)
(4,0.003111)
(8,0.002756)
(16,0.003105)
(32,0.003444)
(64,0.003971)
(128,0.00595)
(256,0.010098)
(512,0.018293)
(1024,0.024934)
(2048,0.058078)
(4096,0.11392)
(8192,0.22557)
(16384,0.5189)
(32768,1.0365)
(65536,1.7795)
(131072,5.8501)
(262144,8.6833)
(524288,18.9347)
};
\addlegendentry{Proposal}

\addplot coordinates {
(2,0.047773)
(4,0.020642)
(8,0.020269)
(16,0.023927)
(32,0.035622)
(64,0.06036)
(128,0.10852)
(256,0.20847)
(512,0.40054)
(1024,0.79939)
(2048,1.6139)
(4096,3.215)
(8192,6.4269)
(16384,12.947)
(32768,26.0436)
(65536,52.1553)
(131072,104.3907)
(262144,209.6472)
(524288,424.8351)
};
\addlegendentry{Wallach}

\end{loglogaxis}
\end{tikzpicture}
\end{center}

While eventually the proposed algorithm performs worse than the Fastfit version, this only occurs when $M$ is close to $10,000$.  In many circumstances, it is not neccesary to have this many samples per row.  If the categorical parameters for that row is already known with such high precision, it might make sense to use the fast Dirichlet solution (section \ref{fastdirichlet}) or subsample the rows to a reasonable size.  Note that as $N$ increases, the lines will cross at higher and higher values of $M$.

The next graph shows the difference between the two phases with ever increasing $M$.  Since the matrix $U$ has $M$ columns, the Newton-Raphson step can no longer remain constant.  However, it still settles at around $80\%$ the speed of the precomputation step as $M$ grows large.

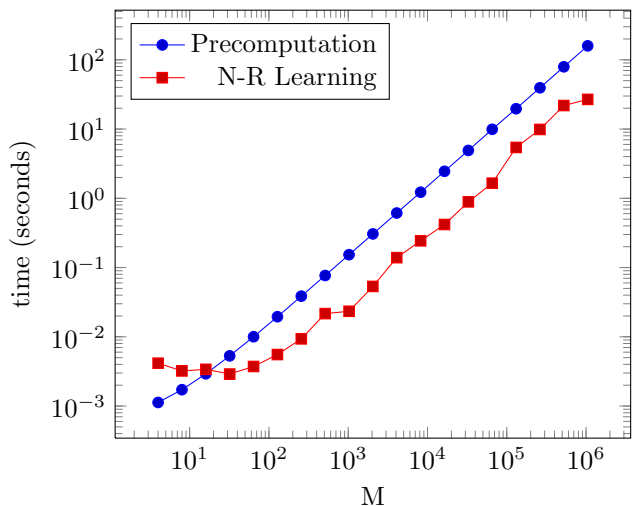
\begin{center}
\begin{tikzpicture}
\begin{loglogaxis}[
xlabel=M,
ylabel=time (seconds),
legend style={
cells={anchor=east},
legend pos= north west,
}]

\addplot coordinates {
(4,0.001127)
(8,0.001723)
(16,0.002921)
(32,0.005306)
(64,0.010026)
(128,0.019507)
(256,0.038626)
(512,0.076808)
(1024,0.15338)
(2048,0.30605)
(4096,0.61295)
(8192,1.2249)
(16384,2.4548)
(32768,4.9129)
(65536,9.9213)
(131072,19.6542)
(262144,39.4747)
(524288,79.2815)
(1048576,159.1461)
};
\addlegendentry{Precomputation}

\addplot coordinates {
(4,0.004161)
(8,0.003213)
(16,0.003392)
(32,0.002902)
(64,0.003728)
(128,0.005545)
(256,0.009336)
(512,0.021624)
(1024,0.023384)
(2048,0.05344)
(4096,0.13907)
(8192,0.24325)
(16384,0.4186)
(32768,0.8884)
(65536,1.6551)
(131072,5.4275)
(262144,9.8792)
(524288,21.8791)
(1048576,26.8208)
};
\addlegendentry{N-R Learning}

\end{loglogaxis}
\end{tikzpicture}
\end{center}

Finally, the algorithms were tested with different values for $K$. The constants were set at $M=50$, $N=5000$, and $\alpha_k = 1/K$.  The benefits of the proposed algorithm are evident through $2048$ dimensions. The negative results for very high $K$ contradict our running-time analysis, and may be caused by issues related to memory allocation and the internal data structures.

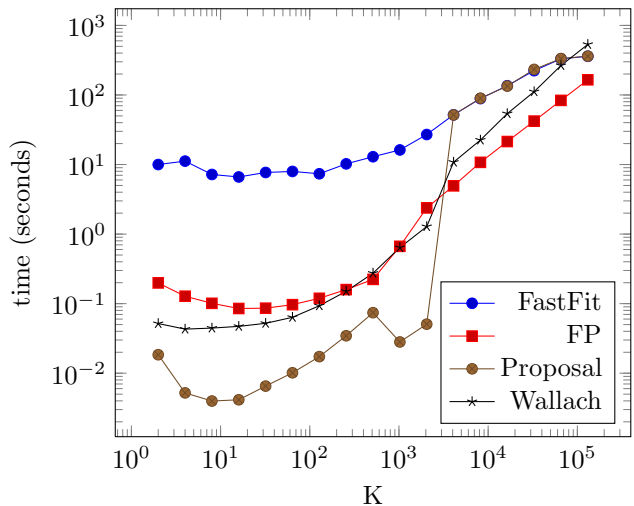
\begin{center}
\begin{tikzpicture}
\begin{loglogaxis}[
xlabel=K,
ylabel=time (seconds),
legend style={
cells={anchor=east},
legend pos= south east,
}]

\addplot coordinates {
(2,9.9911)
(4,11.194)
(8,7.1935)
(16,6.626)
(32,7.6864)
(64,7.9553)
(128,7.3629)
(256,10.2227)
(512,12.9409)
(1024,16.2093)
(2048,26.9865)
(4096,51.8192)
(8192,88.6905)
(16384,135.7325)
(32768,223.4621)
(65536,330.6516)
(131072,359.802)
};
\addlegendentry{FastFit}

\addplot coordinates {
(2,0.19834)
(4,0.12788)
(8,0.10109)
(16,0.085152)
(32,0.086065)
(64,0.096672)
(128,0.11937)
(256,0.15921)
(512,0.22328)
(1024,0.66705)
(2048,2.3781)
(4096,4.9361)
(8192,10.7462)
(16384,21.4049)
(32768,42.1389)
(65536,83.5834)
(131072,165.2163)
};
\addlegendentry{FP}

\addplot coordinates {
(2,0.018433)
(4,0.005211)
(8,0.003981)
(16,0.004149)
(32,0.006524)
(64,0.010139)
(128,0.017307)
(256,0.034532)
(512,0.074105)
(1024,0.028119)
(2048,0.050693)
(4096,51.5911)
(8192,89.7038)
(16384,134.1459)
(32768,233.0731)
(65536,333.6214)
(131072,361.4261)
};
\addlegendentry{Proposal}

\addplot coordinates {
(2,0.05176)
(4,0.043032)
(8,0.044532)
(16,0.047116)
(32,0.052131)
(64,0.063602)
(128,0.093282)
(256,0.15052)
(512,0.27418)
(1024,0.63972)
(2048,1.2835)
(4096,10.8125)
(8192,22.4796)
(16384,53.9609)
(32768,111.5116)
(65536,265.1683)
(131072,533.0268)
};
\addlegendentry{Wallach}

\end{loglogaxis}
\end{tikzpicture}
\end{center}

\section{Conclusions}

Theoretical analysis and experiments suggest that the proposed algorithm for computing the MLE for a Dirichlet-multinomial has improved running times over typical implementations where $M$ is small and $N$ is large.

The proposal also provides a separation between the precomputation step and the Newton-Raphson algorithm.  In the precomputation step, the results from each row can be added independently and therefore it is easy to parallelize, stop early, or add more data.  The Newton-Raphson step, not having to read the entire dataset, runs in constant time as the number of rows increases.

\subsection{Future Work}

In many Dirichlet-multinomial datasets, the amount of data in each row follows a power law distribution.  In other words, a small number of rows contain a large number of samples, and a large number of rows (the \textit{long tail}) contain a small number of samples.  There are two possibilities for handling this case.

First, if the highly sampled rows could be subsampled, the value of $M$ would decrease and the proposed algorithm would run faster. Some analysis would have to be done on exactly how valuable that data is with respect to the MLE for the Dirichlet, but intuitively that would be the least valuable data to sample away.

Another possibility is to pick a desired value for $M$ and split the dataset into two parts.  The first part contains all the rows that have more than $M$ samples, and the second will be used to compute $U$ and $\mathbf{v}$.  Both of these datasets will be far smaller than the original.  Minka's technique will be applied to the first set to find the gradient and Hessian, and the proposed technique will be used for the second set with the results added together.  This hybrid approach would perform better than either of the two approaches independently. The value for $M$ may not even need to be predetermined. A data structure could keep track of the ``free rows'', $U$ and $\mathbf{v}$.  This data structure would accept new rows and only go back and subsume those rows into $U$ and $\mathbf{v}$ when it shrinks the size of the representation.

Another application of this technique is to use it to compute mixture models, which are far more powerful than a single Dirichlet-multinomial model.  In a simple mixture model it is assumed that each row in the dataset was produced by a multinomial coming from a single Dirichlet model in the mixture.  Such a model could favor Dirichlets with high weights so that it looks like a fuzzier version of the multinomial mixture model.  The Latent Dirichlet Allocation is more complex, since different points in each row could be produced by different distributions.  The proposed algorithm, along with the expectation-maximization algorithm, could be used to build these.

Finally, a parallelized map-reduce version of algorithm~\ref{alg:uv} (the precomputation) can be built.  For accurately estimating a Dirichlet-multinomial, a large dataset is overkill.  But suppose that a mixture model is being computed with many Dirichlet clusters to choose from over a sharded dataset.  In this case, a parallelized version of algorithm~\ref{alg:uv} becomes more important.

\subsection{The Code}
\label{code}

Two implementations for the MLE estimate of the Dirichlet-multinomial had been developed in 2014: a Python version and a Matlab version. The Matlab version\footnote{\url{https://github.com/maxsklar/research/tree/master/2014\_05\_Dirichlet/matlab}} extends the fastfit library and was developed for the for the experiments in section~\ref{section:experiments}. The python version is meant for public use and is hosted in github in both an actively developed repository\footnote{\url{https://github.com/maxsklar/BayesPy/tree/master/DirichletEstimate}} and a repository specifically for this paper with both the updated 2023 version\footnote{\url{https://github.com/maxsklar/research/tree/master/2023\_05\_Dirichlet/python}} (now Python 3) and the original 2014 version\footnote{\url{https://github.com/maxsklar/research/tree/master/2014\_05\_Dirichlet/python}}. The Dirichlet MLE estimate relies on NumPy, but the Dirichlet-multinomial uses only functions from the math package. 


\section{Change Notes}
\textbf{May, 2023}
\begin{itemize}
    \item Fixed an error in algorithm~\ref{alg:step} where an incorrect version of the log step was cited without the correct algorithm for the main step. Now, we show both the usual and log step function, and wrote an addendum\cite{addendum} to explain the generation of the algorithm in more detail.
    \item Fixed some minor variable name errors, and tweaked the writing and latex style for clarity and consistency.
    \item Distinguished between the categorical and multinomial distributions. The term \textit{multinomial} had previously been used as a standin for both.
\end{itemize}
\end{document}